\documentclass[runningheads]{llncs}
\usepackage[T1]{fontenc}
\usepackage{multirow}
\usepackage[english]{babel}
\usepackage{graphicx}
\usepackage[colorlinks, linkcolor=black, citecolor=black, urlcolor=black]{hyperref}
\usepackage{geometry}
\usepackage{todonotes} 
\usepackage{ifthen}
\usepackage{listings}
\usepackage[english]{babel} 
\usepackage{amsmath} 
\usepackage{graphicx,wrapfig} 
\usepackage{amssymb}
\usepackage{subcaption} 
\usepackage{xcolor}
\usepackage{float}
\usepackage{bm}
\usepackage{etoolbox}

\usepackage{makecell} 
\usepackage{fullpage} 
\usepackage{hyperref} 
\usepackage{svg} 
\usepackage{epstopdf} 
\usepackage{graphicx}
\begin{document}
\title{On-the-Fly Data Augmentation for Brain Tumor Segmentation}
%
%
\author{Ishika Jain\inst{1,2}\orcidID{0009-0002-8876-5165} \and
Siri Willems\inst{1}\orcidID{0000-0002-4269-1976} \and
Steven Latre\inst{2,1}\orcidID{0000-0003-0351-1714} \and
Tom De Schepper\inst{1}\orcidID{0000-0002-2969-3133}}
\authorrunning{Jain et al.}

\institute{imec, Kapeldreef 75, 3001 Leuven, Belgium \and Department of Computer Science UAntwerp$-$imec, Sint$-$Pietersvliet 7, 2000 Antwerp, Belgium} 

\maketitle              
\begin{abstract}
Robust segmentation across both pre-treatment and post-treatment glioma scans can be helpful for consistent tumor monitoring and treatment planning. BraTS 2025 Task 1 addresses this by challenging models to generalize across varying tumor appearances throughout the treatment timeline. However, training such generalized models requires access to diverse, high-quality annotated data, which is often limited. While data augmentation can alleviate this, storing large volumes of augmented 3D data is computationally expensive. To address these challenges, we propose an on-the-fly augmentation strategy that dynamically inserts synthetic tumors using pretrained generative adversarial networks (GliGANs) during training. We evaluate three nnU-Net-based models and their ensembles: (1) a baseline without external augmentation, (2) a regular on-the-fly augmented model, and (3) a model with customized on-the-fly augmentation. Built upon the nnU-Net framework, our pipeline leverages pretrained GliGAN weights and tumor insertion methods from prior challenge-winning solutions. An ensemble of the three models achieves lesion-wise Dice scores of 0.79 (ET), 0.749 (NETC), 0.872 (RC), 0.825 (SNFH), 0.79 (TC), and 0.88 (WT) on the online BraTS 2025 validation platform. This work ranked first in the BraTS Lighthouse Challenge 2025 Task 1- Adult Glioma Segmentation.

 \and Ensemble 
\end{abstract}
\section{Introduction}
Gliomas are the most common malignant primary brain tumors found in adults. These diffuse gliomas present significant clinical challenges due to their highly infiltrative growth into normal tissue, heterogeneous biology, and variable response to therapy. Accurate assessment of these tumors is critical for diagnosis, treatment planning, monitoring progression and predicting patient outcomes\cite{ref_book2}\cite{cao2023swinunetr}. These assessments are performed by precise delineation, mainly performed manually, of different regions of the tumor, based on information from various MRI modalities (T1,T2, FLAIR, ...). However, manual segmentation of gliomas is time-consuming, subjective and prone to human error, which motivates the need for automated and reproducible solutions.\par 
Over the past 15 years, AI has rapidly evolved, driven by breakthroughs in neural network architectures and learning paradigms. Initially, Convolutional Neural Networks (CNNs) revolutionized image processing, including medical imaging tasks, by excelling at detecting spatial patterns \cite{jia2024application}. Frameworks as U-Net became popular for solving unimodal and multi-modal medical imaging challenges \cite{nnunet}. The introduction of transformers marked a significant shift in AI, enabling models to capture long-range dependencies and contextual relationships within data \cite{vaswani2017attention}. Vision transformers like TransUNet \cite{chen2021transunet}, for example, excel at understanding spatial correlations across the entire images and have significantly improved both the accuracy and efficiency of segmentation methods, outperforming CNNs in scenarios with large, diverse datasets \cite{Dosovitskiy2020vision}. In general, these models leverage large-scale data, self-configuring pipelines, and attention mechanisms to better capture complex anatomical structures, reducing manual effort and accelerating clinical workflows.\par 

To drive research further towards development of robust data-driven clinical pipelines, different challenges tackling various clinical problems are hosted at the annual Medical Image Computing and Computer Assisted Intervention (MICCAI). The Brain Tumor Segmentation (BraTS) challenge in particular provides standardized, annotated datasets and a fair, open benchmarking platform to drive research in automated brain tumor segmentation\cite{bakas2018identifying}. The 2025 BraTS subchallenge on pre- and post-treatment glioma \cite{ref_UBaid}\cite{verdier}\cite{ref_book2}\cite{ref_book3} focuses on robust, automated segmentation of both pre- and post-treatment MRI images specifically, from adults diagnosed with diffuse gliomas. Hereby, aiming to create tools for objectively assessing tumor volume for treatment planning on one hand, and post surgical monitoring and outcome prediction on the other hand.\par

\subsection{State of the Art}\label{sec:sota}
A well-known challenge for solving medical tasks using a data-driven approach is the scarcity of data. While there are millions of images of natural scenes, medical imaging datasets consist of only hundred or couple of thousand samples. Data augmentation strategies are key for the development of robust and strong performing models, which has been proven in previous editions of the BraTS challenge. In the BraTS 2023 \textbf{pre-treatment} adult glioma segmentation challenge, top-performing methods focused on enhancing segmentation precision through architectural innovations and advanced augmentation. The winning team "Faking It" introduced GliGAN, a GAN-based data augmentation framework that inserts realistic synthetic tumors into healthy MRIs using a Swin UNETR-based generator \cite{ferreira2024we}. Other methods employed robust 3D U-Net with attention mechanism \cite{cao2023swinunetr}, and lesion-wise loss function \cite{zhao2023lesionwise} to better capture tumor boundaries and sub-regions. Some achieved competitive results by leveraging model diversity  and robust post-processing \cite{hossain2023advanced}. The winners of the BraTS 2024 \textbf{post-treatment} glioma segmentation challenge were the same team "Faking It", continuing their success by generating massive synthetic data and ensembling diverse architectures trained across multiple folds. Other teams explored ideas, such as generating an additional input image via a linear combination of MRI sequences to emphasize contrast-enhancing tumor regions. This artificial sequence, proposed by Kim \cite{kim2024effective}, improved segmentation accuracy when used alongside the original modalities in ensemble models.\par

While the winning methods in the BraTS 2023 and 2024 challenges achieved notably higher segmentation scores, particularly through advanced augmentation strategies, model ensembling, and synthetic data generation, several limitations remain.

\begin{itemize}
    \item Augmentation using registration took around 2 weeks, ensembling multiple models substantially increases computational cost.

    \item Underrepresented subregions, remain difficult to segment accurately due to low lesion frequency and volume.

    \item In post-treatment gliomas, the presence of multiple small lesions, often low contrast makes accurate segmentation difficult, contributing to lower Dice scores and higher lesion-level false negatives, particularly for the tumor core.
\end{itemize}

In this work, we address these challenges by integrating on-the-fly data augmentation into the nnU-Net training pipeline. This approach not only enables the dynamic insertion of synthetic tumors during training but also allows for targeted augmentation—such as adding small lesions or omitting specific tumor classes—to address class imbalance. Furthermore, we ensemble the augmented models with a baseline model to leverage both the diversity introduced through augmentation and the stability of the original training distribution, aiming for a well-balanced and robust segmentation performance.

\section{Methods}

\subsection{Data}
\begin{figure}[H]
    \centering
    \includegraphics[width = 0.8\textwidth]{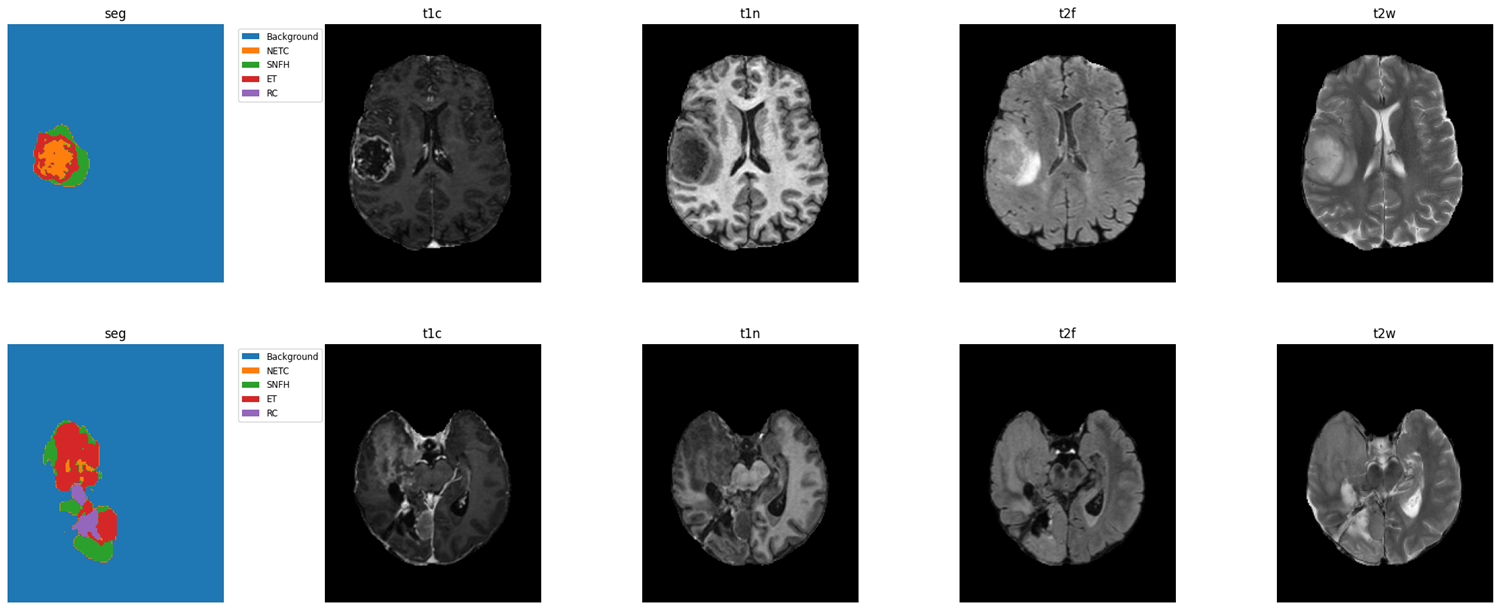}
    \caption{Central axial slice of pre-treatment scans(upper row) and post-treatment scans (lower row)}
    \label{fig:data}
\end{figure}

The data provided by the BraTS 2025 challenge is acquired over multiple institutions and includes routine pre- and post-treatment multi-parametric MRI (mpMRI) from patients diagnosed with diffuse gliomas resulting in total in 2877 cases. Each case in the dataset has four co-registered mpMRI modalities in NIfTI format: native T1-weighted (t1n), post-contrast T1-weighted (t1ce), T2-weighted (t2w), and T2 FLAIR (t2f). All volumes are isotropically resampled to a resolution of 1mm$^3$ with dimensions (182,218,182). Corresponding annotations (Figure \ref{fig:data}) consist of four tumor subregions to which we refer as 'Tumor Classes' from now on:\\
\begin{itemize}
    \item Non-enhancing tumor core (NETC)- the necrosis and cysts within the tumor
    \item Surrounding non-enhancing FLAIR hyperintensity (SNFH)- edema, infiltrating tumor, and post-treatment changes
    \item Enhancing tumor (ET)- regions of active tumor
    \item Resection cavity (RC) -recent and chronic resection cavities (only in post-treatment scans)
\end{itemize}

For clinical decision-making, two composite regions are also of importance and are thus also considered for evaluation:

\begin{itemize}
    \item Tumor core (TC) includes NETC and ET (labels 1 and 3), representing the lesion that is typically removed during surgical resection.

    \item Whole tumor (WT) includes all abnormal tissue (labels 1, 2, 3), representing the whole extent of the tumor
\end{itemize}

The data is divided into a training, validation and test set. The training set includes $\sim2800$ cases (70\%), with annotations available for each case. The validation set includes $\sim400$ cases (10\%), split into 219 pre-treatment and 188 post-treatment subjects. The ground truth annotations are unavailable for the validation set. The test set comprises $\sim800$ cases (20\%), used for final evaluation is unknown.

\subsection{Network architecure}
The nnU-Net framework is a self-configuring deep learning framework designed for medical image segmentation. It has been the core engine behind many winning entries in the BraTS challenges over recent years, including both baselines and customized or ensembled solutions \cite{ferreira2024we}. Therefore, we also used the default fully automated nnU-Net framework\cite{nnunet} (3D full resolution), without any configuration changes. The input is random patches of the shape 128×160×112. Batch size 2, class-based training, and deep supervised.

\subsection{On-the-fly Data augmentation}
 
For this work, we introduce the concept of on-the-fly data augmentation, hereby using GliGan, the GAN developed by previous winners in 2023 and 2024 \cite{ferreira2024we}. The generator of the conditional GAN takes a modality with added noise in healthy areas (where the synthetic tumor has to be inserted) and a label mask as input. The generator replaces the noisy regions with a synthetic tumor that matches the shape and labels defined in the given label mask. The original work generated over 23,000 synthetic scans using this augmentation strategy applying a mix of randomly generated labels and real labels from other subjects\footnote{\href{https://github.com/ShadowTwin41/BraTS_2023_2024_solutions}{BraTS 2023-24 Challenge Winners Solution Github}}.\\

\begin{figure}[H]
    \centering
    \includegraphics[width = 0.8\textwidth]{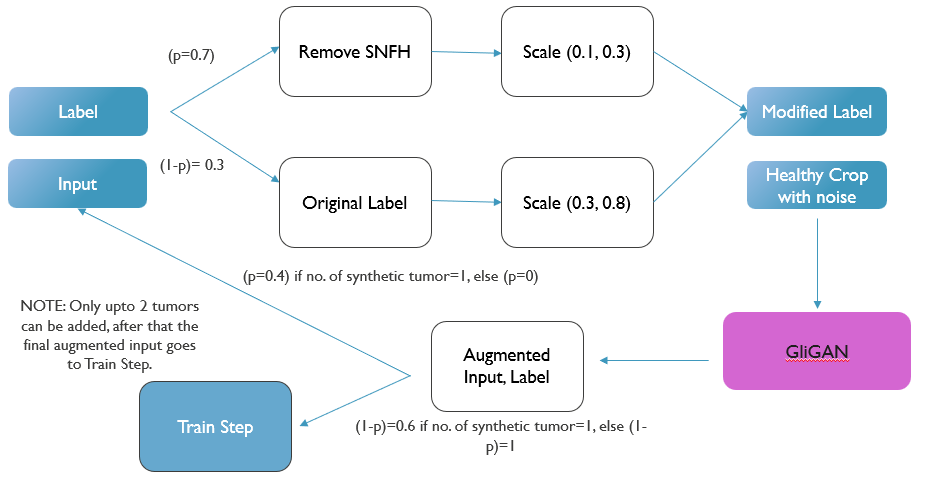}
    \caption{Label modifier framework: if synthetic tumor probability, a randomly selected label mask is modified by probabilistically removing SNFH, scaling, and adding more synthetic tumor(s).}
    \label{fig:label}
\end{figure}

We extend this approach by incorporating on-the-fly GAN-based data augmentation, leveraging the publicly released pre-trained weights of GliGANs for all modalities\footnote{\href{https://doi.org/10.5281/zenodo.14001262}{GliGAN Pre-trained Weights}}. On-the-fly data augmentation refers to the process of applying data transformations dynamically during training, rather than pre-processing and storing augmented data beforehand. These augmentations are performed before the training step for each training batch. The validation data remain unaltered. The following features are added to the existing framework, see Figure \ref{fig:on_the_fly_aug}:
\begin{figure}[H]
    \centering
    \includegraphics[width = 0.8\textwidth]{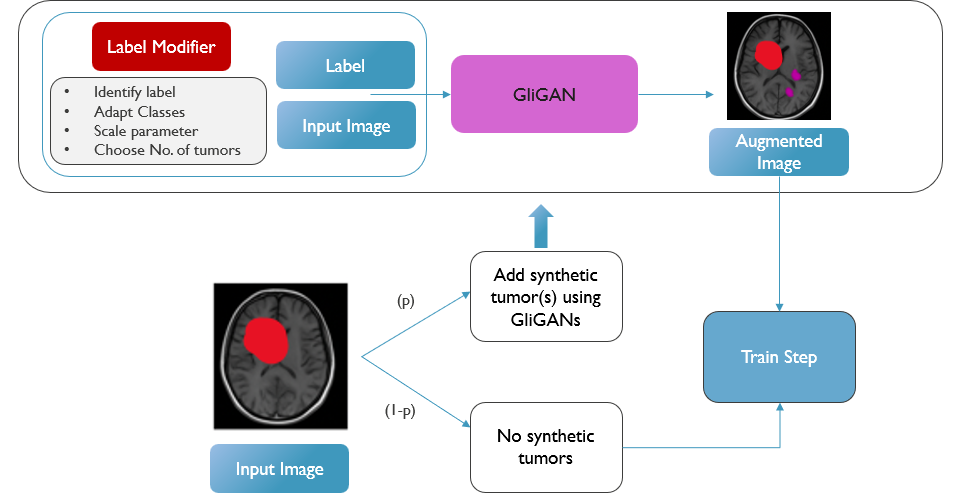}
    \caption{On-the-fly augmentation framework: It adds synthetic tumor(s) to input image with the given probability p before proceeding to the train-step in nnU-Net training pipeline. The GliGAN uses randomly selected label from other patients and the input image to generate synthetic tumor(s) in healthy parts and returns the augmented image and label. The augmented image is now forwarded to the train step. (1-p) times input images remain unmodified.}
    \label{fig:on_the_fly_aug}
\end{figure}
\begin{itemize}
    \item Instead of generating and storing synthetic data beforehand, our pipeline integrates GliGAN-based augmentation directly into the training loop. During each training step, the image is modified with a given probability p to insert synthetic tumor.
    \item Taking advantage of the conditional property of GliGAN, we direct the generator by adapting the input label mask, as illustrated in the zoomed-in view of \textbf{label modifier} in Figure \ref{fig:label}. First we select a label randomly from the remaining patients' labels, then we use the following parameters to modify them:
    \begin{itemize}
        \item Adapt Tumor Classes: To handle the tumor class imbalance issue, the tumor class SNFH is replaced with ET, and ET is then replaced with NETC with a probability 0.7. This increases the number of ET and NETC tumor classes in the augmented data.
        \item Scale: Since nn-UNet model uses the combination of dice scores and cross entropy loss for training, it tends to underperform on small lesions because, they contribute less to overall loss. Moreover, their features are harder to learn due to size and signal-to-noise ratio. Thus, we add the scale parameter which can be used to make the synthetic lesions smaller by simply scaling down the real label. Based on whether SNFH is present, the scale parameter randomly selects a value within a defined upper and lower bound to adapt the synthetic tumor size. If SNFH is removed, the bounds are (0.1,0.3) and if not (0.3, 0.8). The chosen scales are less than 1 because the nnU-Net patch size is relatively small than the input image size. Augmenting smaller lesions increases their frequency, giving the model more chances to learn their patterns.

        \item Choose Number of synthetic tumors to add: Unlike previous approaches that add only one synthetic tumor, our framework supports inserting up to two synthetic tumors per case. This is achieved by introducing an additional probability(0.4), if triggered, repeats the tumor insertion process. The loop continues until a maximum of two synthetic tumors have been added. Once this limit is reached, the final augmented image and corresponding label are forwarded to the training pipeline. This improves lesion-wise sensitivity and better represent multi-lesion scenarios by introducing controlled diversity, while keeping the number of synthetic tumors within a realistic range.
    \end{itemize}
\end{itemize}

This allows for flexible control over augmentation parameters, such as the proportion of augmented samples(p), lesion size (through scaling), and number of synthetic tumors per image. Additionally, it significantly reduces storage requirements, as no intermediate augmented data needs to be saved.

\subsection{Experiments and Evaluation}
We first split the data into training(85\%- 2447 cases) and test(15\%- 430 cases), to have an internal test set to evaluate the performances of models with different modifications.
As a \textbf{Baseline Model (1)}, the default nnU-Net framework and hyperparameters is trained using 5 fold cross-validation using the training data without any external data augmentation using GANs. For the \textbf{On-the-fly Regular Augmentation Model (2)}, the training step of nnU-Net framework was modified to augment the input by adding synthetic tumors in a regular way with the following parameters:
\begin{itemize}
    \item The probability to add synthetic tumor in the input image was set to 0.75. Implying 25\% input images were not modified. 
    \item If selected to augment, the synthetic tumor would be scaled down to a random value selected between (0.3, 0.8).
    \item The probability to add the second synthetic tumor was set to 0, implying that only one synthetic tumor is allowed per input.
\end{itemize}
A 5-fold cross-validation was performed using the training data.

For the \textbf{On-the-fly Custom Augmentation Model (3)}, the training step of nnU-Net framework was modified to augment the input by adding synthetic tumors in customized way with the following parameters:
\begin{itemize}
    \item The probability to add synthetic tumor(s) in the input image was set to 0.6. Implying 40\% input images were not modified. 
    \item If selected to augment, the probability to remove SNFH label was set to 0.7. Thus, 70\% of the augmented lesions will have only the labels ET, NETC, and RC. 
    \item The probability to add the second synthetic tumor was set to 0.4, implying the probability of the input image to have 2 synthetic tumors is (0.6*0.4=0.24) 1 synthetic tumor is (0.6*0.6= 0.36), and no synthetic tumor is (0.4).
\end{itemize}
Due to time constraints, the model was trained once on the full training set instead of performing a 5-fold cross validation. Last, an \textbf{Ensemble} strategy was applied to combine predictions from different models. nnU-Net performs voxel-wise averaging of the softmax outputs across all given models. The final prediction is obtained by taking the argmax over the averaged softmax map. This simple yet effective strategy helps improve segmentation performance by leveraging complementary strengths of different models. We tested all possible combinations of ensemble. 

\begin{itemize}
    \item Baseline + On-the-fly Regular Aug(1+2)
    \item Baseline + On-the-fly Custom Aug(1+3)
    \item On-the-fly Regular Aug + On-the-fly Custom Aug (2+3)
    \item Baseline + On-the-fly Regular Aug + On-the-fly Custom Aug (1+2+3)
\end{itemize}

\textbf{Post Processing:} \\
Since tumor class RC (Resection Cavity) does not occur in pre-treatment cases, any predictions of RC in these cases were adjusted by setting the predicted probability of RC to zero, and then reassigning the label based on the next highest tumor class probability (i.e., using argmax after modification).

\textbf{Thresholding:}\\
Lesion-wise metrics are used for evaluation, meaning the results are particularly sensitive to false positives (FP), i.e. segmentations of non-existent tumors, and false negatives (FN), i.e. missed tumors. While false negatives are difficult to correct through post-processing, false positives can be reduced by applying a threshold to remove small unusual segmentations. The threshold is defined as the minimum number of voxels a predicted lesion must have to be retained in the final segmentation. In particular, post-treatment cases often contain smaller lesions, necessitating more conservative thresholds based on the number of voxels, compared to pre-treatment cases. Even though this may lead to missing some small lesions in pre-treatment cases, it is a trade-off that significantly reduces false positives. For challenge purposes, we use higher thresholds, but in real-world clinical scenarios, thresholding should be avoided to prevent missing smaller lesions, as experts are better positioned to make informed decisions about lesion significance. The following voxel-based thresholds were tested on the best-performing model:

\begin{itemize}
    \item \textbf{Threshold Set 1}: Common threshold for all cases: WT 200, TC 100, ET 60, RC 70
    \item \textbf{Threshold Set 2}: Different threshold for pre-treatment and post-treatment cases:
    \begin{itemize}
        \item Pre-treatment threshold- WT 250, TC 150, ET 100
        \item Post-treatment threshold WT 200, TC 100, ET 50, RC 80
    \end{itemize}
    
\end{itemize}

For evaluation of the above mentioned experiments, the internal test set was used. It allows us to better understand the networks performance and perform smaller ablation studies before submitting anything to the leaderboard platform. This means for each experiment in this study, we show results for an internal validation set first using Dice Similarity Coefficient (DSC) and lesion wise DSC. In addition, the best performing model(s) are submitted to the leaderboard and their performances on the validation and test sets provided by the challenge are used for final evaluation.\\
All experiments were conducted on a high-performance compute node equipped with AMD EPYC 7302 16-Core Processors, 515 GB of RAM, NVIDIA A100 GPU (PCIe, 40GB VRAM).\\

\section{Results}

All the models explained in the above section were evaluated on the internal test set using the threshold 'set 1': WT 200, TC 100, ET 60, RC 70. Table \ref{tab:model_comparison} shows that the scores for all the models are very similar to each other, which makes direct comparison of models challenging. Each model excels in different tumor classes - for instance, Baseline (1) achieved top performance in RC, SNFH, and WT regions, while On-the-fly Regular (2) performed best in TC, making it difficult to identify a single "best" model through simple visual inspection. Thus, to objectively determine the most balanced and consistently performing model we use the following ranking methodology:
We ranked each of the 7 models from 1 (best) to 7 (worst) across all 6 tumor classes based on their lesion dice scores, then calculated average rankings to identify overall performance as seen in Table \ref{tab:lesion_dice_rankings}. This systematic approach revealed Ensemble (1+3) as the top performer with the lowest average rank of 2.67, demonstrating  balanced segmentation across all brain tumor classes.

\begin{table}[H]
\centering
\caption{Performance on Internal Test Set  Comparison: Legacy and Lesion Dice Scores (Mean ± Std)}
\label{tab:model_comparison}
\small
\begin{tabular}{lc|c|c|c|c|c|c}
\hline
\textbf{Metric} & \textbf{Model} & \textbf{ET} & \textbf{NETC} & \textbf{RC} & \textbf{SNFH} & \textbf{TC} & \textbf{WT} \\
\hline
\multirow{7}{*}{\rotatebox{90}{\textbf{Legacy Dice}}} 
& Baseline (1) & 0.829 ± 0.261 & 0.809 ± 0.302 & 0.893 ± 0.236 & 0.876 ± 0.156 & 0.843 ± 0.262 & 0.914 ± 0.129 \\
& On-the-fly Regular (2) & 0.829 ± 0.258 & 0.812 ± 0.294 & 0.885 ± 0.246 & 0.877 ± 0.145 & 0.844 ± 0.259 & 0.915 ± 0.115 \\
& On-the-fly Custom (3) & 0.828 ± 0.256 & 0.825 ± 0.277 & 0.889 ± 0.237 & 0.871 ± 0.147 & 0.842 ± 0.257 & 0.912 ± 0.116 \\
& Ensemble (1+2) & 0.829 ± 0.261 & 0.813 ± 0.295 & 0.894 ± 0.234 & 0.878 ± 0.149 & 0.845 ± 0.259 & 0.916 ± 0.120 \\
& Ensemble (1+3) & 0.833 ± 0.254 & 0.823 ± 0.283 & 0.893 ± 0.233 & 0.877 ± 0.148 & 0.847 ± 0.254 & 0.916 ± 0.118 \\
& Ensemble (2+3) & 0.832 ± 0.254 & 0.818 ± 0.287 & 0.889 ± 0.238 & 0.876 ± 0.146 & 0.846 ± 0.255 & 0.915 ± 0.115 \\
& Ensemble (1+2+3) & 0.829 ± 0.260 & 0.820 ± 0.286 & 0.894 ± 0.234 & 0.878 ± 0.147 & 0.843 ± 0.261 & 0.916 ± 0.117 \\
\hline
\multirow{7}{*}{\rotatebox{90}{\textbf{Lesion Dice}}} 
& Baseline (1) & 0.812 ± 0.267 & 0.821 ± 0.288 & 0.894 ± 0.233 & 0.818 ± 0.213 & 0.825 ± 0.267 & 0.849 ± 0.205 \\
& On-the-fly Regular (2) & 0.813 ± 0.264 & 0.824 ± 0.281 & 0.883 ± 0.246 & 0.815 ± 0.206 & 0.827 ± 0.262 & 0.846 ± 0.202 \\
& On-the-fly Custom (3) & 0.813 ± 0.262 & 0.830 ± 0.271 & 0.888 ± 0.236 & 0.802 ± 0.214 & 0.821 ± 0.265 & 0.835 ± 0.208 \\
& Ensemble (1+2) & 0.812 ± 0.267 & 0.823 ± 0.283 & 0.893 ± 0.233 & 0.817 ± 0.210 & 0.826 ± 0.266 & 0.848 ± 0.204 \\
& Ensemble (1+3) & 0.816 ± 0.262 & 0.835 ± 0.268 & 0.892 ± 0.234 & 0.813 ± 0.213 & 0.827 ± 0.262 & 0.843 ± 0.207 \\
& Ensemble (2+3) & 0.815 ± 0.260 & 0.831 ± 0.272 & 0.889 ± 0.237 & 0.805 ± 0.215 & 0.826 ± 0.263 & 0.837 ± 0.209 \\
& Ensemble (1+2+3) & 0.812 ± 0.267 & 0.833 ± 0.271 & 0.893 ± 0.233 & 0.813 ± 0.210 & 0.822 ± 0.269 & 0.844 ± 0.204 \\
\hline
\end{tabular}
\end{table}

\begin{table}[H]
\centering
\caption{Lesion Dice Performance Rankings (1 = Best, 7 = Worst)}
\label{tab:lesion_dice_rankings}
\small
\begin{tabular}{c|c|c|c|c|c|c|c}
\hline
\textbf{Model} & \textbf{ET} & \textbf{NETC} & \textbf{RC} & \textbf{SNFH} & \textbf{TC} & \textbf{WT} & \textbf{Avg Rank} \\
\hline
Baseline (1) & 5 & 7 & 1 & 1 & 5 & 1 & 3.33 \\
On-the-fly Regular (2) & 3 & 5 & 7 & 3 & 1 & 3 & 3.67 \\
On-the-fly Custom (3) & 3 & 4 & 6 & 7 & 7 & 7 & 5.67 \\
Ensemble (1+2) & 5 & 6 & 2 & 2 & 3 & 2 & 3.33 \\
\textbf{Ensemble (1+3)} & \textbf{1} & \textbf{1} & 4 & 4 & \textbf{1} & 5 & \textbf{2.67} \\
Ensemble (2+3) & 2 & 3 & 4 & 6 & 3 & 6 & 4.0 \\
Ensemble (1+2+3) & 5 & 2 & 2 & 4 & 6 & 4 & 3.83 \\
\hline
\end{tabular}
\end{table}
Table \ref{tab:threshold_analysis} gives an overview of the influence of different thresholds applied to the best performing model on the internal test set, i.e. the  Ensemble (1+3). Set 0 corresponds to No threshold and 1,2 correspond to the respective thresholds mentioned thereafter. The performance improved after applying a thresholding step. Table \ref{tab:threshold_stats} presents the results of paired t-tests conducted between threshold sets. The analysis reveals that both threshold sets 1 and 2 show statistically significant improvements over the no-threshold baseline (set 0), while no significant difference was observed between sets 1 and 2. Based on these findings, we choose to submit both threshold sets for evaluation on the online validation leaderboard.

\begin{table}[h]
\centering
\caption{Threshold Analysis of Ensemble(1+3) Model on Internal Test Set: Legacy and Lesion Dice Scores}
\label{tab:threshold_analysis}
\small
\begin{tabular}{c|c|c|cccc|c|c|c|c|c|c}
\hline
& & & \multicolumn{4}{c|}{\textbf{Thresholds}} & \multicolumn{6}{c}{\textbf{Dice Scores (mean ± std)}} \\
\textbf{Metric} & \textbf{Type} & \textbf{Set} & \textbf{WT} & \textbf{TC} & \textbf{ET} & \textbf{RC} & \textbf{ET} & \textbf{NETC} & \textbf{RC} & \textbf{SNFH} & \textbf{TC} & \textbf{WT} \\ \hline

\multirow{6}{*}{\rotatebox{90}{\textbf{Legacy Dice}}} 
 & Pre & \multirow{2}{*}{0} & 0 & 0 & 0 & 0 & \multirow{2}{*}{0.809 ± 0.279} & \multirow{2}{*}{0.822 ± 0.283} & \multirow{2}{*}{0.868 ± 0.270} & \multirow{2}{*}{0.877 ± 0.148} & \multirow{2}{*}{0.837 ± 0.267} & \multirow{2}{*}{0.916 ± 0.118} \\
 & Post &  & 0 & 0 & 0 & 0 & & & & & & \\ 

 & Pre & \multirow{2}{*}{1} & 200 & 100 & 60 & 70 & \multirow{2}{*}{0.833 ± 0.254} & \multirow{2}{*}{0.823 ± 0.283} & \multirow{2}{*}{0.893 ± 0.233} & \multirow{2}{*}{0.877 ± 0.148} & \multirow{2}{*}{0.847 ± 0.254} & \multirow{2}{*}{0.916 ± 0.118} \\
 & Post &  & 200 & 100 & 60 & 70 & & & & & & \\ 
 & Pre & \multirow{2}{*}{2} & 250 & 150 & 100 & 0 & \multirow{2}{*}{0.830 ± 0.258} & \multirow{2}{*}{0.823 ± 0.281} & \multirow{2}{*}{0.892 ± 0.234} & \multirow{2}{*}{0.877 ± 0.148} & \multirow{2}{*}{0.845 ± 0.257} & \multirow{2}{*}{0.916 ± 0.119} \\
 & Post &  & 200 & 100 & 50 & 80 & & & & & & \\
\hline
\multirow{6}{*}{\rotatebox{90}{\textbf{Lesion Dice}}} 
 & Pre & \multirow{2}{*}{0} & 0 & 0 & 0 & 0 & \multirow{2}{*}{0.792 ± 0.285} & \multirow{2}{*}{0.834 ± 0.269} & \multirow{2}{*}{0.864 ± 0.273} & \multirow{2}{*}{0.806 ± 0.218} & \multirow{2}{*}{0.819 ± 0.271} & \multirow{2}{*}{0.838 ± 0.211} \\
 & Post &  & 0 & 0 & 0 & 0 & & & & & & \\ 

 & Pre & \multirow{2}{*}{1} & 200 & 100 & 60 & 70 & \multirow{2}{*}{0.816 ± 0.262} & \multirow{2}{*}{0.835 ± 0.268} & \multirow{2}{*}{0.892 ± 0.234} & \multirow{2}{*}{0.813 ± 0.213} & \multirow{2}{*}{0.827 ± 0.262} & \multirow{2}{*}{0.843 ± 0.207} \\ 
 & Post &  & 200 & 100 & 60 & 70 & & & & & & \\ 
 & Pre & \multirow{2}{*}{2} & 250 & 150 & 100 & 0 & \multirow{2}{*}{0.811 ± 0.266} & \multirow{2}{*}{0.835 ± 0.268} & \multirow{2}{*}{0.891 ± 0.235} & \multirow{2}{*}{0.813 ± 0.213} & \multirow{2}{*}{0.825 ± 0.264} & \multirow{2}{*}{0.842 ± 0.207} \\
 & Post &  & 200 & 100 & 50 & 80 & & & & & & \\
\hline
\end{tabular}
\end{table}

\begin{table}[H]
\centering
\caption{Statistical Analysis of Lesion Dice Scores Across Threshold Sets (Paired t-tests)}
\label{tab:threshold_stats}
\small
\begin{tabular}{c|c|c|c}
\hline
\textbf{Comparison} & \textbf{t-statistic} & \textbf{p-value} & \textbf{Significant (p < 0.05)} \\
\hline
Threshold Set 0 vs 1 & -2.6963 & 0.0430 & \textbf{Yes} \\
Threshold Set 1 vs 2 &  1.9640 & 0.1067 & No \\
Threshold Set 0 vs 2 & -2.5887 & 0.0489 & \textbf{Yes} \\
\hline
\end{tabular}
\end{table}

\begin{table}[h]
\centering
\caption{Performance of Models on Online Validation Platform: Legacy and Lesion Dice Scores}
\label{tab:model_analysis_val}
\small
\begin{tabular}{lc|c|c|c|c|c|c|c}
\hline
\textbf{Metric} & \textbf{Method} & \textbf{ET} & \textbf{NETC} & \textbf{RC} & \textbf{SNFH} & \textbf{TC} & \textbf{WT} & \textbf{Mean} \\
\hline
\multirow{7}{*}{\rotatebox{90}{\textbf{Legacy Dice}}} 
& Baseline(1)  & 0.799&	0.712	&0.862	&0.869&	0.807	&0.927&	0.829
 \\
& On-the-fly Regular Aug(2)  & 0.801	& 0.713 &	0.853	& 0.868	& 0.810 &	0.925 &  0.828 \\ 

& On-the-fly Custom Aug(3) & 0.793	& 0.712 & 	0.870 &	0.864 &	0.807 &	0.924 &	0.828\\
& Ensemble (1+2) & 0.802 & 0.715 & 0.862 & 0.870 & 0.813 & 0.927 & 0.832 \\
& Ensemble (1+3) & 0.795 & 0.713 & 0.859 & 0.868 & 0.811 & 0.927 & 0.829 \\
& Ensemble (2+3) & 0.795 & 0.710 & 0.866 & 0.867 & 0.810 & 0.926 & 0.829 \\
& Ensemble (1+2+3) & 0.800 & 0.717 & 0.865 & 0.870 & 0.812 & 0.927 & 0.832 \\
\hline
\multirow{7}{*}{\rotatebox{90}{\textbf{Lesion Dice}}} 
& Baseline(1)  & 0.787	& 0.744	& 0.866	&0.825&	0.783&	0.878&	0.814
 \\
& On-the-fly Regular Aug(2) & 0.79 & 0.747 & 0.857 & 0.821 & 0.788 & 0.875 & 0.813 \\
& On-the-fly Custom Aug(3) & 0.783 & 0.745 & 0.878 & 0.814  & 0.784 &	0.871 & 0.812 \\
& Ensemble (1+2) & 0.791 & 0.745 & 0.866 & 0.826 & 0.79 & 0.879 & 0.816 \\
& Ensemble (1+3) & 0.787 & 0.746 & 0.867 & 0.820 & 0.791 & 0.875 & 0.814 \\
& Ensemble (2+3) & 0.786 & 0.742 & 0.874 & 0.821 & 0.788 & 0.876 & 0.814 \\
& Ensemble (1+2+3) & 0.790 & 0.749 & 0.872 & 0.825 & 0.790 & 0.880 & 0.818 \\
\hline
\multirow{7}{*}{\rotatebox{90}{\textbf{Legacy NSD 1.0}}} 
& Baseline(1)  &0.835&	0.733&	0.867&	0.86	&0.784	&0.871	&0.825
\\
& On-the-fly Regular Aug(2)  & 0.839	0.739 & 	0.859	& 0.860 &	0.789 &	0.871 & 0.826\\
& On-the-fly Custom Aug(3) &0.829&	0.734&	0.877&	0.856&	0.784	&0.867 & 0.824
 \\
& Ensemble (1+2) & 0.839 & 0.737 & 0.868 & 0.862 & 0.789 & 0.872 & 0.828 \\
& Ensemble (1+3) & 0.832 & 0.733 & 0.866 & 0.860 & 0.789 & 0.872 & 0.825 \\
& Ensemble (2+3) & 0.833 & 0.732 & 0.872 & 0.860 & 0.788 & 0.871 & 0.826 \\
& Ensemble (1+2+3) & 0.837 & 0.739 & 0.871 & 0.861 & 0.789 & 0.872 & 0.828 \\
\hline
\multirow{7}{*}{\rotatebox{90}{\textbf{Lesion NSD 1.0}}} 
& Baseline(1)  & 0.822	&0.757	&0.871	&0.819 &	0.761 &	0.828	&0.81
 \\
& On-the-fly Regular Aug(2)  & 0.828 & 0.763 & 0.861 & 0.817 & 0.769 & 0.827 & 0.811 \\ 
& On-the-fly Custom Aug(3) & 0.819 &	0.76	& 0.883 &	0.809 &	0.764 &	0.821 & 0.809
	\\
& Ensemble (1+2) & 0.827 & 0.759 & 0.872 & 0.821 & 0.768 & 0.830 & 0.813 \\ 
& Ensemble (1+3) & 0.823 & 0.761 & 0.872 & 0.816 & 0.770 & 0.826 & 0.811 \\
& Ensemble (2+3) & 0.824 & 0.757 & 0.878 & 0.817 & 0.769 & 0.827 & 0.812 \\
& Ensemble (1+2+3) & 0.826 & 0.762 & 0.877 & 0.820 & 0.769 & 0.830 & 0.814 \\
\hline
\end{tabular}
\begin{flushleft}
\centering
\end{flushleft}
\end{table}

\textbf{Challenge Leaderboard}
The validation results on the leaderboard for the above models along with the threshold applied can be seen in Table \ref{tab:model_analysis_val}. This evaluation is performed online as the participants have no access to the ground-truth. The threshold set 2 performed better on validation set than the threshold set 1, thus threshold set 2, i.e. for pre-treatment cases- WT 250, TC 150, ET 100, and for post-treatment cases WT 200, TC 100, ET 50, RC 80 has been applied to all the models. The mean value for lesion-wise metrics is best for the model Ensemble (1+2+3), although other ensembles reach a similar performance.
\section{Discussion }

The BraTS 2025 Challenge – Glioma Segmentation on Pre- and Post-treatment MRI (Task 1) aims for multi-class segmentation of multi-modal MRI scans involving four tumor classes: ET, NETC, RC, and SNFH. The evaluation emphasizes lesion-wise performance, using Dice and normalized surface Dice (NSD) metrics.

In this work, we contribute to the challenge by proposing a segmentation pipeline that combines models trained using on-the-fly data augmentation. Our results demonstrate that ensembling models with and without augmentation—can lead to strong and robust lesion-wise performance using the strengths of each models- realism and diversity.

Both performance on internal test set and the validation leaderboard demonstrate that, overall, individual models and their ensembles produce comparable performance metrics, with each model exhibiting strengths for specific tumor classes. For instance, Baseline Model (1) achieves higher scores for the SNFH tumor class, whereas on-the-fly augmentation Models 2 and 3 show relatively better performance in the TC tumor classes. The ensemble models tend to offer intermediate performance, with a balancing and a trade-off effect: while they reduce the number of false positives by resolving spurious detections present in individual models, they can also eliminate rare true positive cases, marginally increasing false negatives. The ensemble 1+2+3 model performs the best on the lesion-wise scores on the validation platform. 

 Model 3 (the on-the-fly custom augmentation model) had the lowest lesion-wise Dice rankings among the evaluated models, potentially due to its lack of 5-fold cross-validation during training, which may have limited its robustness. Nevertheless, when Model 3 is combined in an ensemble with the baseline model, this pairing yields the highest overall performance, highlighting the complementary strengths of ensemble learning.

Post-processing with thresholding notably reduces false positives and enhances overall results, which is particularly beneficial in a competitive context such as the challenge; however, this approach is not typically advisable for clinical settings due to the potential risk of missing clinically relevant lesions. Nonetheless, for the challenge setting, thresholding results to significant score improvements.

We can conclude that on-the-fly augmentation contributes to improved robustness and stability when used together with the baseline model in an ensemble. However, the current on-the-fly augmentation pipeline is still in its early stages, and its full impact is difficult to assess due to several limitations:

\begin{itemize}
\item None of the models trained reached full convergence at 1000 epochs; they were still improving when training finished. This was due to time constraints of the challenge and the default nn-UNet settings which chose batch size 2, and 1000 epochs. The default nnU-Net configuration is designed to work without expecting external or custom on-the-fly augmentations—its automatic setup assumes only its built-in data augmentation strategies, which are fixed and not dynamically adapted for outside pipelines. As a result, the automatic configuration (including key parameters like patch size, batch size, and max epochs) does not account for additional, on-the-fly augmentations introduced externally. Therefore, to make the best use of our custom on-the-fly augmentation pipeline, we should have optimized these hyperparameters explicitly in the context of this added augmentation, rather than relying on defaults. Training for more epochs or increasing the batch size would allow the models to fully converge, and might better reveal the impact of on-the-fly augmentation.
\item The GliGANs used for data augmentation were adopted from the BraTS 2024 winners and were trained specifically on post-treatment gliomas. Using GliGANs trained on a combined dataset—including both pre- and post-treatment cases—could provide more generalizable and effective augmentations.
\item Several parts of the on-the-fly augmentation pipeline also still need tuning. For example, parameters such as the probabilities and scales for inserting synthetic tumors were selected at random, and could be optimized further for better results. 
\item We could add feedback mechanisms during
 model training, potentially generating synthetic data on the basis of feedback from the segmentation model performance. 
\end{itemize}

Nevertheless, the on-the-fly augmentation approach offers the advantage of reducing both augmentation time and storage needs, while still effectively enhancing the training data with more flexibility.

  

\begin{credits}
\subsubsection{\ackname} This work has received funding from the Flemish Government under the “Onderzoeksprogramma Artificiele Intelligentie (AI) Vlaanderen” programme.

\end{credits}
%
%
%
%

\end{document}